%% file: PaperForReview.tex
\documentclass[10pt,twocolumn,letterpaper]{article}

\usepackage{wacv}              % To produce the CAMERA-READY version
\usepackage{graphicx}
\usepackage{amsmath}
\usepackage{amssymb}
\usepackage{booktabs}
\usepackage{multirow}
\usepackage{bbding}
\usepackage{pifont}
\usepackage{wasysym}

\usepackage[pagebackref,breaklinks,colorlinks]{hyperref}

\usepackage[capitalize]{cleveref}
\crefname{section}{Sec.}{Secs.}
\Crefname{section}{Section}{Sections}
\Crefname{table}{Table}{Tables}
\crefname{table}{Tab.}{Tabs.}

\def\wacvPaperID{1} % *** Enter the WACV Paper ID here
\def\confName{WACV}
\def\confYear{2024}

\begin{document}

\title{Sea You Later: Metadata-Guided Long-Term Re-Identification for UAV-Based Multi-Object Tracking}

\author{Cheng-Yen Yang${^1}$\thanks{~ Corresponding author: \textit{cycyang@uw.edu}.} \quad Hsiang-Wei Huang$^1$\quad Zhongyu Jiang$^1$\quad Heng-Cheng Kuo$^2$\\ Jie Mei$^1$\quad Chung-I Huang$^3$\quad Jenq-Neng Hwang$^1$\\
[2mm]
$^1$Information Processing Lab, University of Washington, USA \quad
$^2$National Taiwan University, Taiwan\\
$^3$National Center for High-performance Computing, Taiwan\\
% {\tt\small \{cycyang, hwhuang, zyjiang, jiemei, hwang\}@uw.edu\\}
}
\maketitle

%%%%%%%%% ABSTRACT
\begin{abstract}
   Re-identification (ReID) in multi-object tracking (MOT) for UAVs in maritime computer vision has been challenging for several reasons. More specifically, short-term re-identification (ReID) is difficult due to the nature of the characteristics of small targets and the sudden movement of the drone's gimbal. Long-term ReID suffers from the lack of useful appearance diversity. In response to these challenges, we present an adaptable motion-based MOT algorithm, called Metadata Guided MOT (MG-MOT). This algorithm effectively merges short-term tracking data into coherent long-term tracks, harnessing crucial metadata from UAVs, including GPS position, drone altitude, and camera orientations.  Extensive experiments are conducted to validate the efficacy of our MOT algorithm. Utilizing the challenging SeaDroneSee tracking dataset, which encompasses the aforementioned scenarios, we achieve a much-improved performance in the latest edition of the UAV-based Maritime Object Tracking Challenge with a state-of-the-art HOTA of $69.5\%$ and an IDF1 of $85.9\%$ on the testing split.
\end{abstract}

%%%%%%%%% BODY TEXT
\section{Introduction}
\label{sec:intro}

In recent years, the field of computer vision has witnessed remarkable advancements in multi-object tracking (MOT) techniques. These advancements have enabled the development of systems capable of detecting and following objects in various scenarios. However, one challenging and pressing domain where MOT capabilities are sought is maritime computer vision, including a wide range of applications \cite{marques2015mariapplication0, lygouras2019mariapplication1, acharya2021mariapplication2}. The unique characteristics of this environment, including the presence of small-sized objects, challenging visibility conditions due to waves and sun reflections, and the dynamic nature of objects caused by gimbal movements and altitude changes, pose formidable challenges for conventional MOT algorithms. To exacerbate these difficulties, partial occlusions frequently occur in maritime scenes. Addressing these challenges requires a holistic approach that not only detects and tracks objects but also ensures the long-term re-identification of targets that temporarily vanish and reappear. 

% \begin{figure}[t]
%     \centering
%     \includegraphics[width=0.8\linewidth]{fig1_intro_temp.png}
%     \caption{\color{red} PLACEHOLDER for introduction visualization.}
%     \label{fig:intro}
% \end{figure}

\begin{figure*}[t]
    \centering
    \includegraphics[width=\linewidth]{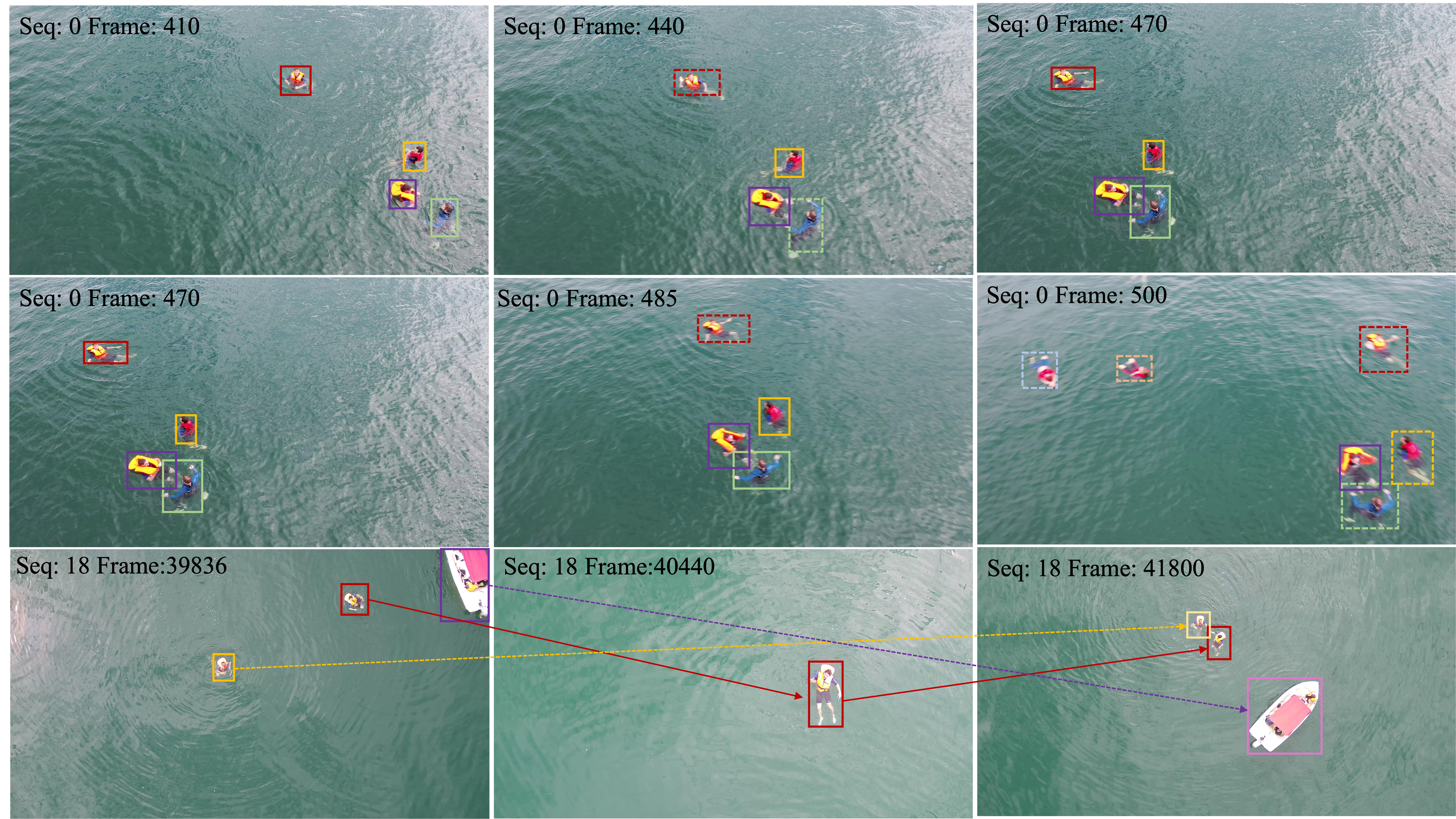}
    \caption{Three types of detection and tracking errors caused by fast drone movement (Top), rapid camera gimbal changes (Mid), and target re-entries (Bottom). Dashed bounding boxes represent the missing detections in our baseline model, while the dashed lines across image frames represent the IDs due to not being able to associate correctly.}
    \label{fig:challengin}
\end{figure*}

% In this paper, we present a groundbreaking solution that leverages drone metadata to facilitate multi-object tracking and tackle the intricate task of long-term re-identification within the SeaDronesSee-MOT benchmark, recently introduced at WACV 2024.

The SeaDronesSee-MOT benchmark \cite{varga2022seadronessee, kiefer20221stmacvi} is specifically designed to assess the capabilities of computer vision algorithms in the maritime domain, emphasizing the detection and tracking of humans, boats, and other objects in open water. While several MOT benchmarks exist, SeaDronesSee-MOT introduces the novel aspect of long-term tracking, a challenge that requires the re-identification (ReID) of objects that temporarily disappear from the scene and subsequently reappear within the same video clip. This challenge is particularly demanding for objects such as boats and swimmers, which may share similar visual characteristics. To address this complex task, we exploit the wealth of drone metadata accompanying each frame, including altitude, viewing angles, and gimbal information, among others. These metadata serve as a valuable resource for accurately associating objects over time, offering a promising avenue for enhancing the robustness and effectiveness of maritime MOT systems. In this paper, we detail our innovative approach, highlighting how the integration of drone metadata enables us to not only excel in multi-object tracking but also excel in long-term ReID, marking a significant step forward in the realm of maritime search and rescue missions.

There are two main challenges for the UAV-based tracking for the maritime:

1) First, the tracking performance is highly dependent on detection qualities, therefore, the nature of the objects in maritime tracking serves as one of the challenges. Similar to the scenario in UAV detection, the scale of the objects we try to detect is highly variant due to the height of the UAVs. Therefore, it is important to reconsider whether using one uniform detector is enough or not.

2) Secondly, long-term and short-term ReIDs are also challenging. For short-term ReID, the difficulties come from the quick movement of either the drone or the camera, such as quick rotating of pitch or yaw can result in unsatisfying bounding box tracking results. In the case of long-term ReID, which is a significant focus of the SeaDronesSee-MOT benchmark, traditional appearance features, often effective in pedestrian tracking scenarios, may not perform well for object tracking in maritime environments. The challenge arises from the characteristics of maritime objects, including boats, which may share similar visual appearances. 

Therefore, the paper argues that leveraging drone metadata, such as altitude, viewing angles, and gimbal information, can help overcome these challenges and significantly enhance the robustness and effectiveness of maritime MOT systems. The integration of the metadata marks a significant step forward in the field of maritime search and rescue missions, enabling more accurate and reliable tracking and ReID of objects over extended time frames. Our proposed Metadata-Guided MOT (MG-MOT) effectively merges short-term tracking data into coherent long-term tracks, harnessing crucial metadata from UAVs, including GPS position, drone altitude, and camera orientations, which achieves first place in the UAV-based Multi-Object Tracking with Re-Identification Track in the latest edition of Workshop on Maritime Computer Vision (MaCVi). 

% Our contributions can be summarized in two folds:

% \begin{itemize}
%     \item Our proposed method...
%     \item We ranked 1st...
% \end{itemize}

The paper is organized as follows: we will first introduce some related prior works towards MOT in Sec \ref{sec:related}. Then our main proposed method used in the challenge will be described in Sec \ref{sec:method}. Sec \ref{sec:dataset} and \ref{sec:exp} will cover the implementation details and the experiments. Finally, we will have the conclusion in Sec \ref{sec:conclusion}.

\section{Related Work}
\label{sec:related}

\noindent\textbf{Multi-Object Tracking.}
Multi-object tracking algorithms have been significantly improved by the advancement in deep learning-based detectors. The recent tracking algorithms usually follow the tracking by detection paradigm and utilize an object detector and an association algorithm to conduct tracking. Several popular tracking algorithms include DeepSORT\cite{wojke2017deepsort}, ByteTrack\cite{zhang2022bytetrack}, and BoTSORT\cite{aharon2022botsort}. These methods usually focus on the tracking of pedestrians and vehicles, in which the target objects and cameras usually demonstrate simple movement. Several popular existing MOT datasets include MOT\cite{MOT16} and BDD\cite{yu2020bdd100k}. However, with the recent increased popularity of maritime computer vision applications, more and more research starts to focus on the MOT task in maritime environments \cite{kiefer20221stmacvi,yang2023multi}.

\noindent\textbf{Tracking with Moving Cameras.}
Strong camera movement can cause failure in object motion modeling and object detection during multi-object tracking tasks. Camera movements exist in multiple benchmarks\cite{MOT16,cui2023sportsmot,kiefer20221stmacvi}. To increase the robustness of tracking and reduce the negative effect of camera motion, BoTSORT \cite{aharon2022botsort} uses the global motion compensation (GMC) technique, allowing the tracker to estimate the background motion and thus produce a more accurate object motion prediction for the association progress. StrongSORT \cite{du2023strongsort} incorporates an enhanced correlation coefficient maximization (ECC) model \cite{evangelidis2008parametric} for camera motion compensation and helps the tracker's estimatation of the global rotation as well as the translation of adjacent frames. While some other algorithms leverage different association methods to conduct tracking under large camera motion, e.g., bounding box distance \cite{huang2023observation}, different IoU association method \cite{huang2023iterative}. These methods aim to generate higher spatial similarity of the same object in different time stamps even if they share no IoU in adjacent frames, thus providing more robustness during the tracking process.

\noindent\textbf{Tracking with Metadata.}
Several previous works leverage the information from metadata to conduct multi-object tracking. The metadata might include useful information like the target object's information, camera-related information (e.g., intrinsic and extrinsic parameters), etc. \cite{hsu2021multi,huang2023multi} utilizes vehicle metadata and vehicle travel distance to increase the ReID accuracy and multi-camera vehicle tracking performance in urban surveillance scenarios. Huang et al. \cite{Huang_2023_aicity} estimates the camera calibration by selecting corresponding points in the bird-eye-view map and camera frames using the PnP method in \cite{Tang18AIC} to further improves the association between camera-views in an indoor scene. Kiefer et al. \cite{kiefer2023memory} show that metadata from UAVs can create a memory map of object locations in actual world coordinates, improving the representation of object locations, which has proven to enhance the downstream tasks, e.g., object detection, multi-object tracking, and video anomaly detection.

\section{Proposed MG-MOT Method}
\label{sec:method}

Due to the nature of the characteristics of small and similar targets and the sudden movement of the drone’s gimbal, appearance-based and 2D motion-based ReID or tracking methods can easily fail in this maritime MOT task. Taking advantage of rich metadata from the drone, including latitude, longitude, altitude, pitch and yaw angles, we are able to construct the camera model and build our long-term ReID method based on 3D geometry. 

\subsection{Estimated Camera Model}
\label{subsection:camera}

% We choose N$47.6^\circ$ E$9.2^\circ$ as the root point of our world coordinate.

\begin{figure}[tb!]
    \centering
    \includegraphics[width=1\linewidth]{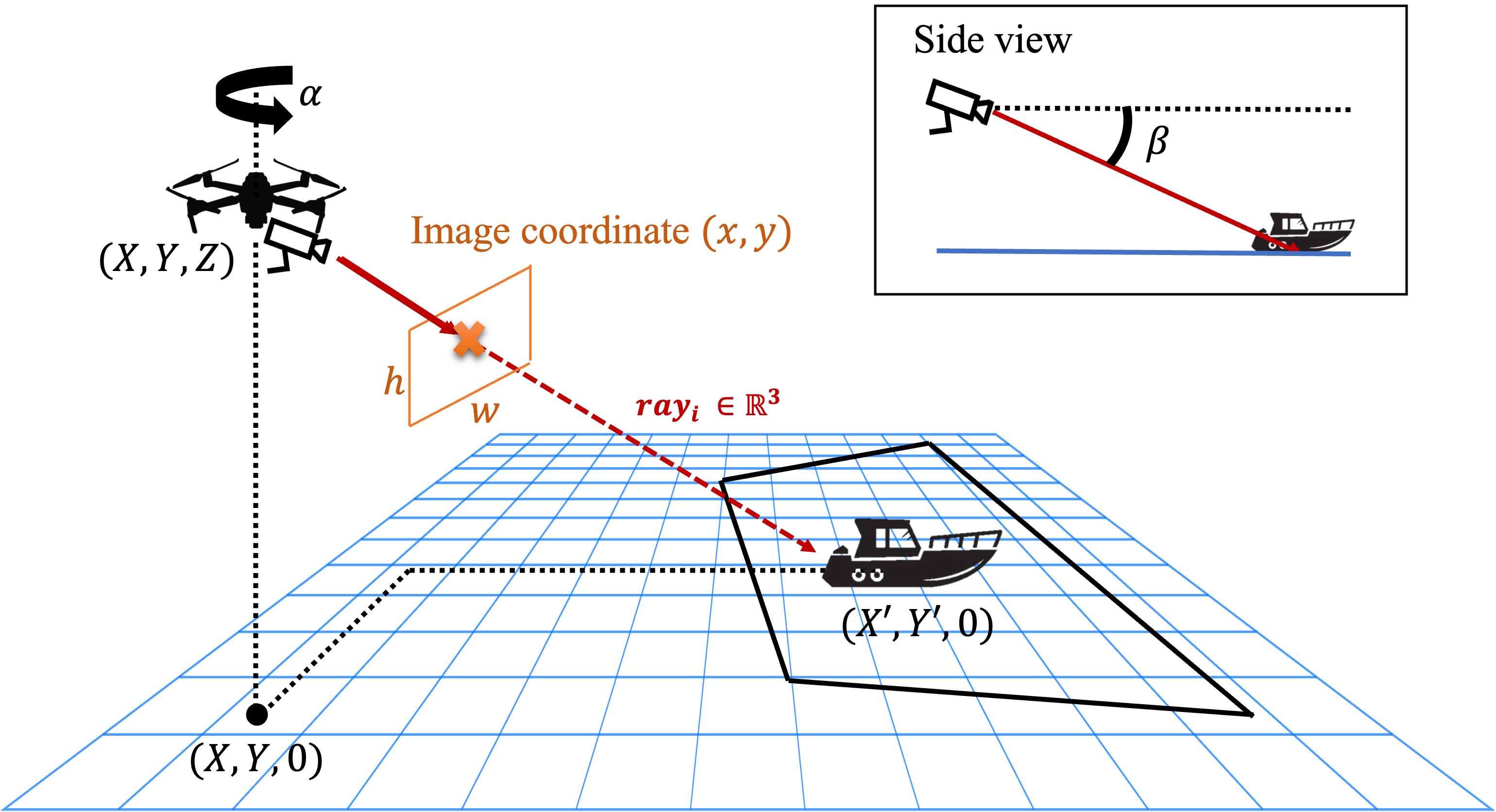}
    \caption{Illustration of our estimated camera model used to project the location of the object from image coordinates $(x,y)$ to world coordinates $(X',Y', 0)$.}
    \label{fig:camera_model}
\end{figure}

More specifically, the camera's intrinsic parameters can be obtained by field-of-view as

\begin{equation}
    K = \begin{bmatrix}
        w/ 2\tan{(fov/2)} & 0 & w/2 \\
        0 & w/2\tan{(fov/2)} & h/2 \\
        0 & 0 & 0
        \end{bmatrix},
\end{equation}

\noindent where $w$ and $h$ are the width and height of images, and $fov$ is the field-of-view of the camera. 

As shown in Fig~\ref{fig:camera_model}, for each frame, with the metadata from the drone, the rotation matrix between the camera coordinate and the world coordinate can be established by:

\begin{align}
    R(\alpha, \beta, 0) &= \begin{bmatrix}
        \cos{\alpha}\cos{\beta} & -\sin{\alpha} & \cos{\alpha}\sin{\beta} \\
        \sin{\alpha}\cos{\beta} & \cos{\alpha} & \sin{\alpha}\sin{\beta} \\
        -\sin{\beta} & 0 & \cos{\beta}
        \end{bmatrix},
\end{align}

\noindent where $\alpha$ and $\beta$ are the yaw and pitch angles of the drone gimbal. Specifically, $\alpha = 90^{\circ} - \textit{gimbal\_heading}$ and $\beta = \textit{gimbal\_pitch}$, where gimbal\_heading and gimbal\_pitch are given by metadata. Combining $K$ and $R$, any detected object $X^{2D}_{i}$ and its corresponding image coordinates $x_i$ can be projected to the world coordinates using the 3D directional vector:

\begin{equation}
    ray_i = R\cdot K^{-1}\cdot \frac{x_i}{||x_i||} \in \mathbb{R}^3.
\end{equation}

Finally, we use the altitude information $Z_{drone}$ to compute the intersection of our $ray_i$ with the sea surface, which we assume is a plane of $z=0$:

\begin{equation}
    X^{3D}_{i} = Loc_{object} = Loc_{drone} + \frac{Z_{drone}}{Z_{ray_i}} \cdot ray_i.
\end{equation}

\subsection{Metadata-Guided Re-Identification}

\noindent \textbf{Short-Term ReID.} For short-term ReID, we focus on efficiently associating the broken tracklets which do not exit the camera view, a scenario that typically occurs within a shorter time window. This often results from sudden drone movements or rapid gimbal adjustments. Given the limited time frame in such situations, we adopt a direct approach by computing the world coordinates directly. This method allows us to track more effectively and enhance our ability to maintain continuous object tracking even during these dynamic and challenging conditions.

For each detection at frame $t$, the world coordinate $X^{3d}_{t}$ is computed using the projection $H(\cdot)$ derived from the metadata $m_t$ as mentioned in Sec \ref{subsection:camera}:

\begin{equation}
    X^{3D}_{t} = H_{m_t}\big( \hat{X}^{2D}_{t} \big),
\end{equation}

\noindent where $\hat{X}^{2D}$ represents the center point of the bounding box $X^{2D}$. For short-term ReID, we directly use Hungarian assignment to match the tracks using the world coordinate distance:
 
\begin{equation}
    \text{arg min } C_{\text{dis}}(T_i, T_j) = \left\| X^{3D}_{t^{\text{exit}}_i} - X^{3D}_{t^{\text{entry}}_j} \right\|_2
\end{equation}

\noindent where $T_i$ represents the tracks left in the memory bank waiting to be associated, while $T_j$ represents the entering or new tracks waiting to be matched. An additional $\tau_{match}$ is used to constraint the matching; if the cost is greater than such threshold, we will treat the entering or new tracks as a new one.

\noindent \textbf{Long-term Re-ID.} Different from short-term ReID, long-term ReID is somewhat more challenging because we need to closely monitor the potential movement of tracked objects which are out of the image view. Therefore, on top of the standard world coordinate distance matching, we add in two crucial components, i.e. \textbf{Bi-directional Movement Extrapolation (BiME)} and \textbf{Matching Threshold Expansion (MTE)}. Bi-Directional Movement Extrapolation is a naive method that extrapolates the world coordinates of the tracks that exit or enter the image. Given some tracks $T_i=\{X_{t^{enter}_{i}},\dots,X_{t^{exit}_{i}}\}$ that enter and exit the image during frame $t^{enter}$ and $t^{exit}$. The world coordinate of $T_i$ at $t'$ will be forward-extrapolating as:

\begin{equation}
\label{eq:extra}
    T^{3D}_i = X^{3D}_{t^{exit}_{i}} + \Delta t \cdot V^{exit}_{T_i} \text{,\ \ \  if }\Delta t < \tau_{memory},
\end{equation}

\noindent where $\Delta t = t' - t^{exit}$ and the track last-seen exit velocity can be estimated as $V_{T_i}$: 

\begin{equation}
\label{eq:velo}
    V^{exit}_{T_i} = \Big( \frac{X^{3D}_{t^{exit}_{i}}-X^{3D}_{t^{exit}_{i}-w}}{w} \Big).
\end{equation}

\input{table/table_dataset}

\input{table/table_metadata}

\noindent using the window size $w$ as a constant. Note that we will not extrapolate further beyond a certain duration $\tau_{memory}$ since the world coordinates after such duration are often unreliable and may be confused with some new tracks. We also backward-extrapolating the tracks using similar logistics in Eq. \ref{eq:extra} and \ref{eq:velo} by substituting the exit terms with the entry terms. In contrast to short-term re-identification, Matching Threshold Expansion is a strategy to expand the matching space as the tracks disappear or reappear from the image:

\begin{equation}
    \tau'_{match} = \lambda \cdot ( \Delta t_{exit} + \Delta t_{entry})  \cdot \tau_{match}.
\end{equation}

\noindent \textbf{Class-wise Re-ID.}
Recognizing the distinct characteristics of boat and swimmer movement patterns and appearances, we conduct short-term and long-term Re-identification (Re-ID) in a class-specific manner. We tailor the thresholds mentioned earlier for each class, such as setting a larger $\tau_{match}$ for boats due to their potential for higher relative velocity. The association steps are performed individually, enabling us to significantly lower the chances of associating tracks from different classes.

\input{table/table_leaderboard}

% \noindent \textbf{Overall ReID Procedure.} We summarize our re-identification into the pseudo-code.

\section{Dataset: SeaDroneSee-MOT}
\label{sec:dataset}

The SeaDronesSee-MOT dataset consists of 21 clips in the training set, 17 clips in the validation set, and 19 clips in the testing set with a total of 54,105 frames and 403,192 annotated instances as in Table \ref{table:dataset}. Metadata for the drone are also provided for all training, validation, and testing split as we summarized it in Table \ref{table:metadata}. Detail studies and analyses on the characteristics of each sequence (e.g., challenging scenario, per sequence HOTA performance, distribution of metadata) are being carefully investigated and reported in \cite{kiefer20221stmacvi}. Note that the training and validation sets do not contain long-term tracking labels, i.e., objects that have gone missing are assigned new IDs when they reappear.

\section{Experiment Results}
\label{sec:exp}

\subsection{Implementation Details}
\label{sec:implementation}

\noindent \textbf{Detector.} We use the YOLOv8-x \cite{yolov8_ultralytics}, pretrained weights on COCO \cite{lin2015coco} and with an additional p2 head, to predict tiny object. We also find the need to train a multi-class detector instead of a single-class one since the characteristics of the boat and swimmer classes are totally different in terms of movement, appearance, and others. We experiment with several settings of input image size along with the input data source (e.g., jpg or png) and record the detection results. All detectors are trained using similar hyper-parameters for 100 epochs with an initial learning rate of $0.01$ and a decay of $0.05$. Eventually, we settle for the image size of $1280$. Our detector performs poorly when detecting swimmers while the drone flew low to the sea surface with a considerable heading angle. This is due to a lack of similar data in the training dataset. Therefore, we adopt a pre-trained YOLOv8-x detector to perform an ensemble of detections for the sake in the challenge. The yolov8-p2 detectors are trained and inferenced using Nvidia Tesla V100 GPUs. 

\noindent \textbf{Tracking.} Before applying the proposed meta-guided ReID method, we obtained the initial tracking results using BoT-SORT \cite{aharon2022botsort} with sparse optical flow as the Generalized Motion Compensation (GMC) method. We use the same tracking hyper-parameters throughout all 18 testing sequences for a generic tracking method. The thresholding for the high confidence detection is set to $0.5$, while the low confidence detection is set to $0.1$. We initialize a new track with a confidence higher than $0.2$ if the detection does not match any existing track and use a buffer of $100$ frames to remove those unmatched tracks.

\noindent \textbf{Camera Parameters Calibration.} We manually pick several segments of sequences from training and validation, with non-moving targeted objects and a continuously moving or turning camera. Then we try to minimize the standard deviation of the locations of the same target with different field-of-view angles. Finally, we select a field-of-view of $70\deg$. The intrinsic matrix is obtained using such field-of-view angle and image size being $3840\times2160$ to obtain the world coordinates of each object from image coordinates.

% \noindent \textbf{Re-Identification.} There are several thresholds we adapt in our metadata-guided re-identification step, including the initial cost, tracklet memory time.

% The reasoning behind having two sets of 

\noindent \textbf{Post-Processing.} We employ a simple linear interpolation strategy to recover missing detections, whether caused by poor lighting conditions or sudden camera movements. We have two iterations of interpolations, one before and one after the ReID step. Additionally, in some sequences with lower altitudes, we may encounter overlapping detections, leading to a degradation in detection performance. To address this issue, we apply non-maximum suppression to filter out overlapping detections. It is important to note that, unlike traditional non-maximum-suppression, we prioritize detections with larger bounding box areas while filtering out smaller boxes, as in most situations, the latter are false positive detections.

\subsection{Results on SeaDroneSee-MOT}

\noindent \textbf{Evaluation Metrics.} The results of long-term multi-object tracking are evaluated using HOTA \cite{luiten2021hota} and CLEAR MOT metrics \cite{CLEARMOT} including MOTA, IDF1, MOTP, MT, ML, FP, FN, Recall, Precision, ID Switches, Fragments. Note that despite the multi-classes labels of the tracks being available in the training and validation data, the testing only considers all tracks as the same classes during evaluation.

\begin{figure}[tb!]
    \centering
    \includegraphics[width=\linewidth]{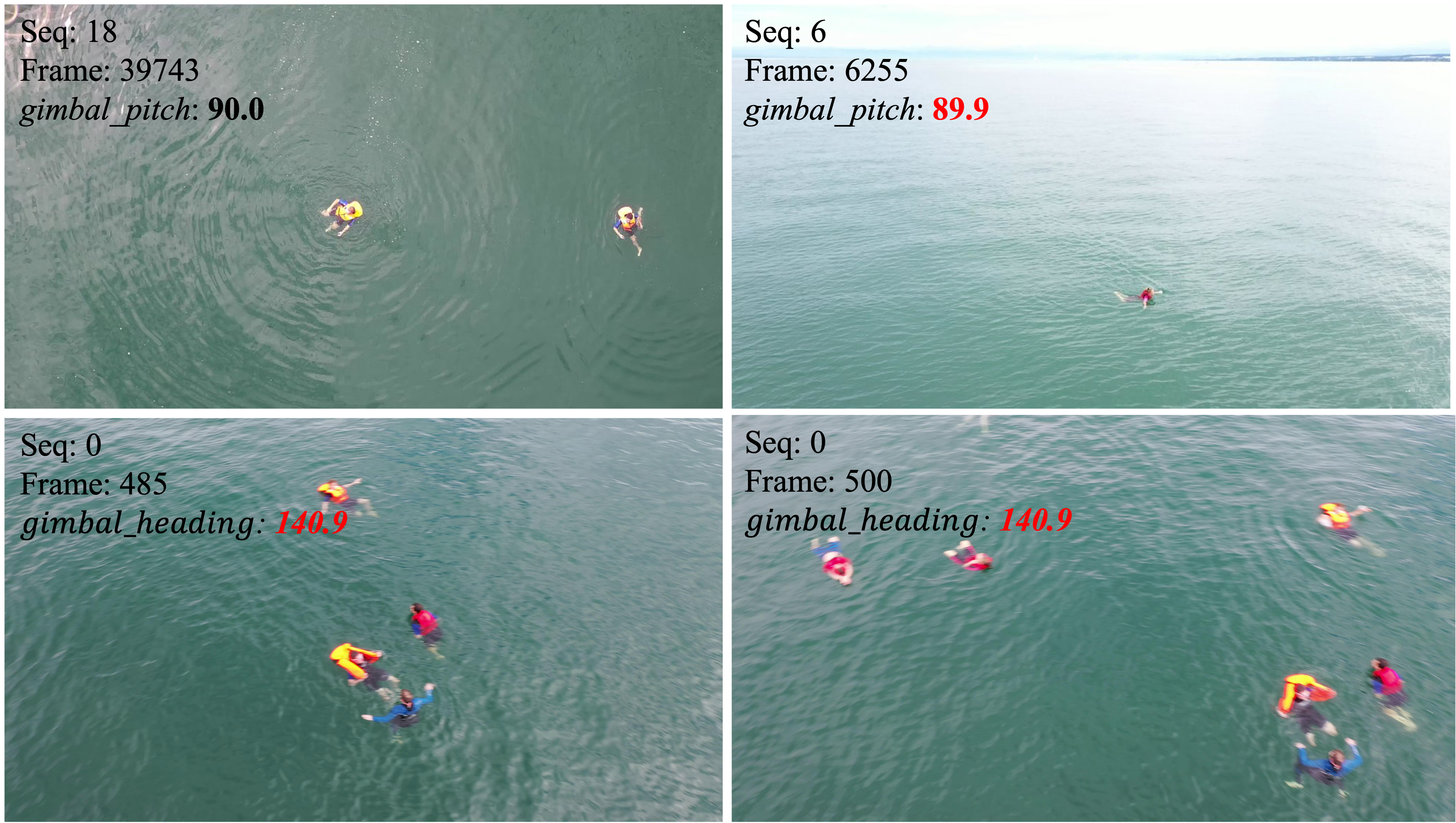}
    \vspace{-1em}
    \caption{Two types of error in our final results. Top row: Example of gimbal pitch error in the metadata found in the testing split. The two frames have nearly identical pitch, but the drone's directions are clearly different. Bottom row: Example of the metadata is not synchronized with the actual drone gimbal.}
    \label{fig:metadata_error}
\end{figure}

\begin{figure}[tb!]
    \centering
    \includegraphics[width=\linewidth]{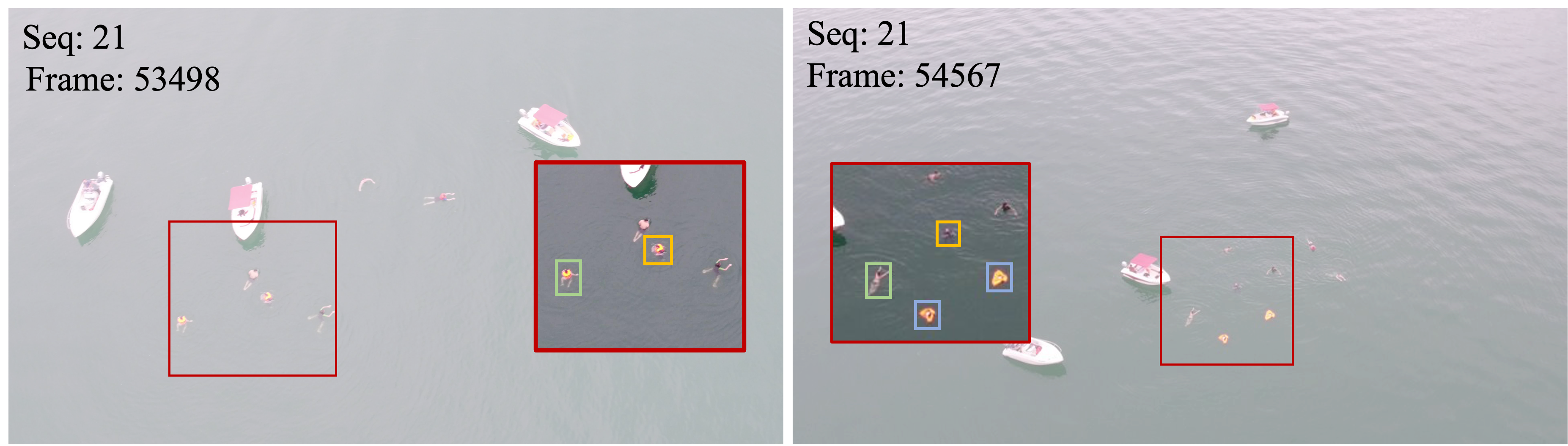}
    \caption{An example of splitted tracks, in the earlier frame, the swimmers are wearing lifejackets, however, in the later frame, after the lifejackets are taken off, new track ids emerges.}
    \label{fig:track_error}
\end{figure}

\noindent \textbf{Final Results.} In the results section of the competition, as illustrated in Table \ref{table:leaderboard}, the performance of various methods and teams in the 2024 MacVi SeaDronesSee Multi-Object Tracking with Reidentification Challenge is rigorously evaluated across several critical metrics (see https://macvi.org/leaderboard/airborne/seadronessee/multi-object-tracking-reid). Notably, our proposed MG-MOT (Team 198) demonstrates exceptional tracking accuracy, achieving the highest HOTA score of $69.5\%$, while also securing the highest IDF1 score, an impressive $85.9\%$. This method's performance showcases the ability to track objects in challenging scenarios accurately. Furthermore, Franunfofer IOSB (Team 220) emerges as a noteworthy contender, securing the highest MOTA at $78.0\%$, highlighting their prowess in the detection results as they also have the fewest FN detection. These results underline the significant advancements in maritime MOT capabilities and affirm the strong potential of our proposed MG-MOT method for real-world applications in object tracking.

\subsection{Ablation Studies}

\begin{figure*}[t]
    \centering
    \includegraphics[width=1\linewidth]{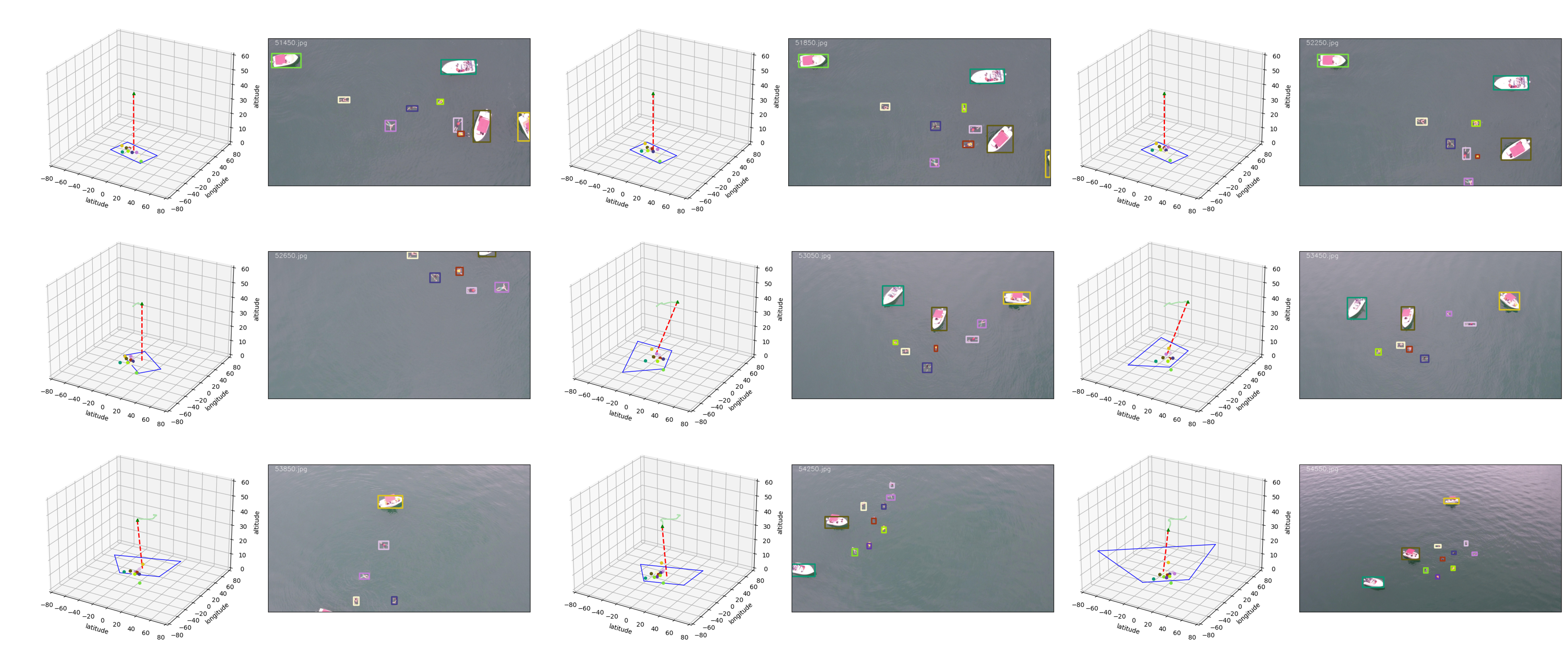}
    \caption{Visualization of the result of MG-MOT on the testing sequence 21 with world coordinates (best view in color).}
    \label{fig:challengin}
\end{figure*}

\input{table/table_detector}
\input{table/table_tracking_ablation}

\noindent \textbf{Re-ID Modules.} The effectiveness of our MG-MOT components, short- and long-term re-identification are being analyzes in Table \ref{table:reid}. Our MG-MOT is compared to the baseline using the BoT-SORT, which also serves as our initial tracking results (in Sec \ref{sec:implementation}). We found that both the short- and long-term re-identification strategies can improve the IDF1 of the tracking therefore boosting the HOTA performance based on the same detections.

\noindent \textbf{Detection.} We report the detection result using different YOLOv8 backbones, input sources, and input sizes in Table \ref{table:detector} with the same tracking hyper-parameters. The performance is obtained by submitting the results to the evaluation server. Due to a significant number of tiny detections in the dataset, the $p2$ head combined with a greater input size of the backbone model resulted in better MOTA, FP, and FN.

\noindent \textbf{Tracking Parameters.} We conduct ablation studies on various tracking parameters' impact in the inference stage. To assess performance on annotated data, a new model is trained based on the same architecture as the previously reported best one, using the training set and evaluating on the validation set. Notably, interpolation and ReID are not adopted. The effect of the IoU threshold on tracking performance, tuning it from 0.1 to 0.9. Results in Table~\ref{table:thresholds} show that setting the IoU threshold to 0.5 achieves a more balanced performance. Detecting swimmers prove to be more challenging than detecting boats due to small bounding boxes and interactions, resulting in box overlaps. The impact of the matching threshold by varying it from 0.1 to 0.9. Results illustrate that a higher matching threshold improves performance. In videos with multiple swimmers, they tend to gather as a group and follow similar trajectories, making a higher matching threshold valuable in preventing incorrect matches.

\subsection{Limitations}

We observe certain limitations in our proposed method and the standard algorithm for UAV-based MOT. One notable challenge is ensuring the synchronization and accuracy of metadata with the captured frames. However, this issue could potentially be resolved by cross-referencing with the presence of the sea surface in the images. Moreover, when objects temporarily vanish from the camera frame due to drone movement or camera repositioning, there is a risk of long-term ReID failure, as these objects may reappear after an unusually prolonged interval. For this, we can employ an expansion of the search and rematch zone or simply readopt the appearance-based ReID for further filtering of possible matches.

\noindent\textbf{Error in Metadata.} We find the metadata of the drone is not that reliable. As shown in Fig~\ref{fig:metadata_error}, although the gimbal pitch angles are almost the same, the image from Seq $6$ shows an entirely different view, which is more similar to the pitch angle around $45^{\circ}$.

\noindent\textbf{Error in Splitted Tracks.} In the testing set, there are some cases when a swimmer takes off his/her life jacket during the tracking process, which causes a "split tracks" situation to happen. Since our tracking algorithm does not consider object class during the association stage, this might result in ID switch during the tracking process and lead to a degradation in tracking performance.

\section{Conclusion}
\label{sec:conclusion}

In summary, our motion-based multi-object tracking algorithm, MG-MOT, enhanced by the inclusion of UAV metadata, represents a significant performance stride in the domain of maritime computer vision. We've made substantial progress by addressing the challenges of short-term and long-term ReID with the help of UAV metadata. The effectiveness of our algorithm, as demonstrated through the SeaDroneSee dataset and the UAV-based Maritime Multi-Object Tracking Challenge, is a testament to its practicality. We achieve a much-improved performance in the latest edition of the UAV-based Maritime Object Tracking Challenge with a state-of-the-art HOTA of 69.5\% and an IDF1 of 85.9\% on the testing split.

\section{Acknowledgment} We want to thank the National Center for High-performance Computing (NCHC) from Taiwan for providing computational and storage resources.

%Future direction of UAV-based MOT for maritime computer vision may include multi-camera or multi-drone systems.

%%%%%%%%% REFERENCES
{\small
\bibliographystyle{ieee_fullname}
\bibliography{egbib}
}

\end{document}

%% file: table/table_dataset.tex
\begin{table}[tb]
\small
\centering
\begin{tabular}{l|cccc}
\hline
Dataset    & \# of Seq. & \# of Frame & \# of Bbox & Longest Seq. \\ \hline
Training   & 20         & 27,259      & 160,470    & 6,296        \\
Validation & 17         & 8,584       & 47,678     & 2,069        \\
Testing    & 19         & 18,253      & -          & 4,138        \\ \hline
Total      & 21         & 54,096      & 207,938    & -            \\ \hline
\end{tabular}
\caption{The overall statistics of the SeaDroneSee-MOT dataset.}
\label{table:dataset}
\end{table}

%% file: table/table_metadata.tex
% "gps_latitude": 47.671949, "gps_latitude_ref": "N", "gps_longitude": 9.269725, "gps_longitude_ref": "E", "altitude": 8.599580615665955, "gimbal_pitch": 45.4, "compass_heading": 138.2, "gimbal_heading": 140.9, "speed": 0.7349802860974518, "xspeed": -0.299991953509164, "yspeed": 0.299991953509164, "zspeed": 0.599983907018328}}

\begin{table}[tb]
\small
\centering
\begin{tabular}{ll|ll}
\hline
\multicolumn{2}{c|}{\multirow{2}{*}{Metadata}}                               & \multicolumn{2}{c}{Data Split (Mean/Min/Max)}             \\ \cline{3-4} 
\multicolumn{2}{c|}{}                                                        & \multicolumn{1}{c|}{Train+Val} & \multicolumn{1}{c}{Test} \\ \hline
\multicolumn{1}{l|}{\multirow{3}{*}{\textbf{P}}} & \textit{gps\_latitude}   & \multicolumn{1}{c|}{0.072/0.070/0.074}     & \multicolumn{1}{c}{0.072/0.071/0.074}                 
\\\multicolumn{1}{l|}{}                             & \textit{gps\_longitude}  & \multicolumn{1}{c|}{0.069/0.067/0.071}          &    \multicolumn{1}{c}{0.069/0.067/0.071}                      \\
\multicolumn{1}{l|}{}                             & \textit{altitude}        & \multicolumn{1}{c|}{34.29/4.90/140.39}          &   \multicolumn{1}{c}{42.55/4.60/149.49}                       \\ \hline
\multicolumn{1}{l|}{\multirow{2}{*}{\textbf{D}}} & \textit{gimbal\_pitch}   & \multicolumn{1}{c|}{42.5/-2.5/90.0}          &     \multicolumn{1}{c}{57.333/6.8/90.0}                     \\
\multicolumn{1}{l|}{}                             & \textit{gimbal\_heading} & \multicolumn{1}{c|}{201.6/0.1/359.8}          &   \multicolumn{1}{c}{184.437/0/359.3}                       \\ \hline
\multicolumn{1}{l|}{\multirow{3}{*}{\textbf{S}}} & \textit{x\_speed}        & \multicolumn{1}{c|}{1.24/0/13.799}          &        \multicolumn{1}{c}{1.331/0/10.799}                  \\
\multicolumn{1}{l|}{}                             & \textit{y\_speed}        & \multicolumn{1}{c|}{0.953/0/12.799}          &      \multicolumn{1}{c}{1.046/0/9.399}                    \\
\multicolumn{1}{c|}{}                             & \textit{z\_speed}        & \multicolumn{1}{c|}{0.129/0/3.999}          &   \multicolumn{1}{c}{0.251/0/5.199}            \\ \hline
\end{tabular}
\caption{The frame-level drone metadata provided in the SeaDroneSee-MOT dataset and its corresponding statistics according to the train/val/test splits. \textbf{P} stands for positions (latitude and longitude are reported with a relative center at (N$47.6^\circ$ E$9.2^\circ$)), \textbf{D} stands for directions, and \textbf{S} stands for absolute speed. }
\label{table:metadata}
\end{table}

%% file: table/table_leaderboard.tex
\begin{table*}[tbh!]
    \centering
    \footnotesize
    \begin{tabular}{l|ccc|ccccccccc}
        \hline
        Team & HOTA$\uparrow$ & MOTA$\uparrow$ & IDF1$\uparrow$ & MOTP$\downarrow$ & MT$\uparrow$ & ML$\downarrow$ & FP$\downarrow$ & FN$\downarrow$ & Rec$\uparrow$ & Pre$\uparrow$ & IDs$\downarrow$ & Frag$\downarrow$ \\ \hline
        Baseline (MaCVi) & 37.5 & 43.8 & 38.8 & 22.9 & 69 & 75 & 13566 & 38858 & 59.4 & 80.7 & 1340 & 2556 \\
        Team 400 & 	43.4 & 53.2 & 46.9 & 22.4 & 86 & 64 & 10587 & 33075 & 65.4 & 85.5 & 1110 & 2233 \\
        NCKU ACVLab (Team 403) & 49.9 & 32.0 & 57.9 & -1.000 & 86 & 97 & 23601 & 41253 & 56.9 & 69.7 & 161 & 646 \\
        MI-SIT (Team 399) & 	53.1 & 62.5 & 58.8 & \textbf{19.8} & 116 & 54 & \textbf{9068} & 26650 & 72.1 & 88.4 & 123 & \underline{896}\\
        Team 412 & 	55.4 & 65.3 & 61.5 & 20.9 & 134 & 31 & 13603 & 19426 & 79.7 & 84.9 & 137 & 1162 \\
        Lenovo (Team 395) &  61.5 & 76.8 & 70.4 & 20.6 & 145 & 31 & 10155 & 11982 & 87.5 & \underline{89.2} & 100 & 1350\\
        Franunhofer IOSB (Team 220) & \underline{69.3} & \textbf{78.0} & \underline{84.4} & \underline{20.5} & \textbf{165} & \textbf{20} & 10643 & \textbf{10391} & \textbf{89.1} & 88.9 & \textbf{16} & 984\\ \hline
        % Ours (baseline) & 0.603 & 0.776 & 0.682 & 0.209 & 159 & 21 & 9932 & 11381 & 0.881 & 0.895 & 84 & 753  \\
        % Ours (baseline + ensemble) & 0.612 & 0.777 & 0.699 & 0.207 & 158 & 22 & 9956 & 11338 & 0.881 & 0.894 & 78 & 769\\
        % Ours (w/ MetaData) & 0.690 & 0.772 & 0.854 & 0.207 & 157 & 22 & 10082 & 11674 & 0.878 & 0.893 & 26 & 809 \\
        % Ours (w/ MetaData + ensemble)  & 0.694 & 0.779 & 0.859 & 0.207 & 157 & 22 & 9626 & 11476 & 0.880 & 0.897 & 20 & 787 \\ 
        Ours (Team 198) & \textbf{69.5} & \underline{77.9} & \textbf{85.9} & 20.7& \underline{158} & \underline{22} &\underline{9700}&\underline{11425}&\underline{88.1}&\textbf{89.7}&\underline{18}&\underline{784}\\\hline
    \end{tabular}
    \caption{Leaderboard Results of the 2024 MacVi SeaDronesSee Multi-Object Tracking with Re-Identification. The \textbf{bold} numbers indicate the best while the \underline{underlined} numbers indicate the second to the best among all participants. Our proposed method achieved the SOTA in terms of HOTA and IDF1 on the testing split with a noticeable amount of FP, ID switches, and track fragments being reduced.}
    \label{table:leaderboard}
\end{table*}

\begin{table*}[tbh!]
    \centering
    \footnotesize
    \begin{tabular}{l|cc|ccc|ccccccccc}
        \hline
        Method & Ensemble & Re-ID & HOTA$\uparrow$ & MOTA$\uparrow$ & IDF1$\uparrow$ & MOTP$\downarrow$ & FP$\downarrow$ & FN$\downarrow$ & Rec$\uparrow$ & Pre$\uparrow$ & IDs$\downarrow$ & Frag$\downarrow$ \\ \hline

         Baseline & & & 60.3 & 77.6 & 68.2 & 20.9  & 9932 & 11381 & 88.1 & 89.5 & 84 & 753  \\
         (BoT-SORT) & \Checkmark & & 61.2 & 77.7 & 69.9 & 20.7  & 9956 & 11338 & 88.1 & 89.4 & 78 & 769\\\hline
         &  & Short & 64.0 &	76.8 & 76.6 & 20.8 & 10543 & 11621 & 87.9 & 88.9 & 	84 & 758  \\
        Ours &  & Long & 69.0 & 77.2 & 85.4 & 20.7  & 10082 & 11674 & 87.8 & 89.3 & 26 & 809 \\
        (MG-MOT) & \Checkmark & Long & 69.4 & \textbf{77.9} & \textbf{85.9} & 20.7  & 9626 & 11476 & 88.0 & 89.7 & 20 & 787 \\ 
         & \Checkmark & Short+Long & \textbf{69.5} & \textbf{77.9} & 85.0 & 20.7 & 9700 & 11425 & 88.1 & 89.7 & 18 & 784\\\hline
    \end{tabular}
    \caption{Our implemented method using different types of detector backbones and different Re-ID method combinations.}
    \label{table:reid}
\end{table*}
% \begin{table*}[tbh!]
%     \centering
%     \footnotesize
%     \begin{tabular}{l|cc|ccc|ccccccccc}
%         \hline
%         Method & Ensemble & Re-ID & HOTA$\uparrow$ & MOTA$\uparrow$ & IDF1$\uparrow$ & MOTP$\downarrow$ & MT$\uparrow$ & ML$\downarrow$ & FP$\downarrow$ & FN$\downarrow$ & Rec$\uparrow$ & Pre$\uparrow$ & IDs$\downarrow$ & Frag$\downarrow$ \\ \hline

%          Baseline & & & 0.603 & 0.776 & 0.682 & 0.209 & 159 & 21 & 9932 & 11381 & 0.881 & 0.895 & 84 & 753  \\
%          (BoT-SORT) & \Checkmark & & 0.612 & 0.777 & 0.699 & 0.207 & 158 & 22 & 9956 & 11338 & 0.881 & 0.894 & 78 & 769\\\hline
%          &  & Long & 0.690 & 0.772 & 0.854 & 0.207 & 157 & 22 & 10082 & 11674 & 0.878 & 0.893 & 26 & 809 \\
%         MG-MOT & \Checkmark & Long & 0.694 & 0.779 & 0.859 & 0.207 & 157 & 22 & 9626 & 11476 & 0.880 & 0.897 & 20 & 787 \\ 
%          & \Checkmark & Short+Long & 0.695 & 0.779 & 0.85} & 0.207& 158& 22&9700&11425&0.881&0.897&18&784\\\hline
%     \end{tabular}
%     \caption{Our implemented method using different type of detector backbones and different Re-ID method combinations. S reprsents short-term and L represents long-term.}
%     \label{tab:my_label}
% \end{table*}

%% file: table/table_detector.tex
\begin{table}[tb!]
\centering
\footnotesize
\begin{tabular}{l|cc|ccc}
\hline
Backbone   & Source & Input Size & MOTA$\uparrow$  & FP$\downarrow$    & FN$\downarrow$    \\ \hline
yolov8x    & jpg    & 640        &  64.3  & 8998      &  25063     \\
yolov8x-p2 & jpg    & 640        &  68.0  & \textbf{8010}     &  22448     \\
yolov8x-p2 & jpg    & 960        &  70.4     &  8971     & 19249      \\
yolov8x-p2 & png    & 960        & 76.0 & 10052 & 12817 \\
yolov8x-p2 & png    & 1280       & \textbf{77.3} & 9701  & \textbf{11930} \\ \hline
\end{tabular}
\caption{Ablation studies on the detector performance.}
\label{table:detector}
\end{table}

%% file: table/table_tracking_ablation.tex
\begin{table*}[tbh!]
\centering
\small
\label{tab:thresholds}
\begin{tabular}{l|c|ccccccccc}
\toprule
Tracking Parameter & Metric & 0.1 & 0.2 & 0.3 & 0.4 & 0.5 & 0.6 & 0.7 & 0.8 & 0.9 \\
\midrule
\multirow{5}{*}{IoU Threshold} 
& HOTA$\uparrow$ & 69.4 & 69.4 & 69.6 & 69.7 & 69.9 & 69.6 & 69.6 & 69.1 & 67.3 \\
& MOTA$\uparrow$ & 80.3 & 80.6 & 81.4 & 81.3 & 81.3 & 81.4 & 81.4 & 81.1 & 80.5 \\
& IDF1$\uparrow$ & 80.8 & 80.7 & 80.6 & 80.5 & 81.0 & 80.6 & 80.6 & 79.2 & 75.7 \\
& FP$\downarrow$ & 7078  & 7262  & 7732  & 7575  & 7340  & 7732  & 7732  & 8225  & 9998  \\
& FN$\downarrow$ & 10573 & 10526 & 10282 & 10467 & 10225 & 10282 & 10282 & 11005 & 12529 \\
\midrule
\multirow{5}{*}{Matching Threshold} 
& HOTA$\uparrow$ & 33.6 & 47.5 & 69.6 & 62.4 & 66.1 & 69.6 & 69.6 & 69.6 & 69.9 \\
& MOTA$\uparrow$ & 21.4 & 55.0 & 81.4 & 77.9 & 80.3 & 81.4 & 81.4 & 81.4 & 81.4 \\
& IDF1$\uparrow$ & 27.6 & 41.2 & 80.6 & 66.9 & 72.2 & 80.6 & 80.6 & 80.6 & 81.4 \\
& FP$\downarrow$ & 7917  & 7917  & 7732  & 7917  & 7917  & 7732  & 7732  & 6769  & 4953  \\
& FN$\downarrow$ & 38711 & 31167 & 10282 & 17455 & 14575 & 10282 & 10282 & 10282 & 9894  \\
\bottomrule
\end{tabular}
\caption{Ablation studies on IoU threshold and matching threshold.}
\label{table:thresholds}
\end{table*}